# COEFFICIENTS' SETTINGS IN PARTICLE SWARM OPTIMIZATION: INSIGHT AND GUIDELINES

**Mauro S. Innocente[a], Johann Sienz[b,c]**

[a]*Research Assistant, Civil & Computational Engineering Centre, College of Engineering, Swansea University, Singleton Park, Swansea (SA2 8PP), Wales, UK, M.S.Innocente@swansea.ac.uk, http://www.swansea.ac.uk/engineering/Research/*

[b]*Professor, Aerospace Research Theme Leader, College of Engineering, Swansea University, Singleton Park, Swansea (SA2 8PP), Wales, UK, J.Sienz@swansea.ac.uk, http://www.swansea.ac.uk/engineering/ http://www.swansea.ac.uk/staff/academic/Engineering/sienzhans/*

[c]*Co-Director of the Welsh Composites Centre (WCC), http://www.welshcomposites.co.uk/*

**Abstract**. Particle Swam Optimization (PSO) is a population-based and gradient-free optimization method developed by mimicking social behaviour observed in nature. Its ability to optimize is not specifically implemented but emerges in the global level from local interactions. In its canonical version, there are three factors that govern a given particle's trajectory: 1) the *inertia from its previous displacement*; 2) the *attraction to its own best experience*; and 3) the *attraction to a given neighbour's best experience*. The importance given to each of these factors is regulated by three coefficients: 1) the *inertia*; 2) the *individuality*; and 3) the *sociality weights*. The settings and relative settings of these coefficients rule the trajectory of the particle when pulled by these two attractors. While divergent trajectories are of course to be avoided, different speeds and forms of convergence of a given particle towards its attractor(s) take place for different settings of the coefficients. A more general formulation is presented, aiming for a better control of the embedded randomness. Guidelines as to how to select the settings of the coefficients to obtain the desired behaviour are offered. As to the convergence speed of the whole algorithm, it also depends on the speed of spread of information within the swarm. The latter is governed by the structure of the neighbourhood, whose study is beyond the scope of the research presented here. The objective of this paper is to help understand the core of the PSO paradigm from the bottom up by offering some insight into the form of the particles' trajectories, and to provide some guidelines as to how to decide upon the settings of the coefficients in the particles' velocity update equation in the proposed formulation to obtain the type of behaviour desired for the given particular problem. General-purpose settings are also suggested, which provide some trade-off between the reluctance to getting trapped in suboptimal solutions and the ability to carry out a fine-grain search. The relationship between the proposed formulation and both the classical and constricted PSO formulations are also provided.

# 1  INTRODUCTION

Particle Swarm Optimization (PSO) is a global optimizer in the sense that it is able to escape poor suboptimal solutions. This is possible thanks to a parallel search carried out by a population of cooperative individuals –called particles– which profit from sharing information acquired through experience.

Individually, each particle is pulled by two attractors, while also carrying some inertia from its previous displacement. One of the attractors is its own best previous experience, and the other is the best previous experience of a given neighbour. Thus, these are the three basic ingredients ruling a particle's trajectory: the *inertia from its previous displacement*; the *attraction to its own best previous experience*; and the *attraction to a given neighbour's best previous experience*. In the classical formulation, the importance granted to each of these three ingredients is controlled by three coefficients: the *inertia*; the *individuality*; and the *sociality weights*. Thus, the individual behaviour of a particle is governed by the settings of these coefficients. Random weights embedded in the particles' velocity update equation introduce creativity into the system so as to avoid getting trapped in some regular pattern.

The other important aspect with regards to the particles' behaviour is which neighbours are to inform of their best experiences to which particles. In other words, how to define the social attractor in the particles' velocity update equation, thus governing the particles' social behaviour. This leads to the development of infinite designs of social networks within the so-called *swarm* (population), which are typically referred to as *neighbourhood structures* or *neighbourhood topologies*.

There is always the need of a trade-off between the explorative and the exploitative behaviour of the particles in the swarm. Explorative behaviour is better in avoiding premature convergence and in escaping local attractors, whereas exploitative behaviour is better in performing a fine-grain search while exhibiting faster convergence. This trade-off may be controlled by both the coefficients' settings and the neighbourhoods' topology. This paper presents a study of the former, whilst the study of neighbourhood topologies is beyond the scope of the research presented here.

The remainder of this paper is organized as follows: the PSO method is reviewed in section 2; the research body of the paper is presented in section 3, where the theoretical convergence studies are offered in section 3.1; different speeds and forms of convergence for different regions of the convergence graph are shown in section 3.2; a reformulation of the PSO basic equation so as to control the range of randomness is proposed in section 3.3; while guidelines for the settings of the coefficients in order to obtain the desired behaviour are provided in section 3.4. Final remarks and lines for future research are presented in section 4.

# 2  PARTICLE SWARM OPTIMIZATION

Particle Swarm Optimization is a population-based and gradient-free optimization method introduced by social-psychologist James Kennedy and electric engineer Russell C. Eberhart in 1995 (Kennedy and Eberhart, 1995). The method was inspired by earlier bird-flock simulations (e.g. Heppner and Grenander (1990); Reynolds (1987)) and strongly influenced by Evolutionary Algorithms (EAs). Therefore it has roots on different fields such as social psychology, Artificial Intelligence (AI), and mathematical optimization. At present, its main applications are in solving optimization problems that are difficult to be handled by traditional methods. The algorithm is especially suitable for nonlinear problems with real-valued variables, although adaptations can be found in the literature to deal with discrete problems (e.g. Kennedy and Eberhart (1997); Kennedy and Eberhart (2001) (pp. 289–299); Mohan and

Al-Kazemi (2001); and Clerc (2004)). Given that gradient information is not required, non-differentiable and even discontinuous problems can be handled. In fact, since the method imposes no restrictions to the functions involved, they do not even need to be explicit.

Since PSO is not deterministically implemented to optimize but to simulate some social behaviour, its optimization ability is an emergent property resulting from local interactions among the particles. This makes it difficult to understand its theoretical bases. Nonetheless, considerable theoretical work has been carried out on simplified versions of the algorithm (e.g. Ozcan and Mohan (1998); Ozcan and Mohan (1999); Kennedy and Eberhart (2001); van den Bergh (2001); Clerc and Kennedy (2002); Trelea (2003); Clerc (2008); Kennedy (2008); and Innocente (2010)). For a comprehensive review of the method, refer to Engelbrecht (2005) and Clerc (2006a). For a short review, see Bratton and Kennedy (2007).

Theoretical studies of the PSO algorithm's behaviour in the presence of randomness are beyond the scope of this thesis. Only few researchers –to the best of our knowledge– dared take this challenge. Jiang et al. (2007) studied the convergence of an isolated particle using stochastic process theory, viewing the particles' position as a stochastic vector. By studying the convergence of the expectation and of the variance of the particle's position, they claim to have derived the 'stochastic convergent condition' of the particle swarm system. Clerc (2006b) studied the stagnation phenomenon in PSO (no improvement observed over several time-steps). In that extensive formal study, he analyzed the distribution of velocities of a particle with stochastic forces. In turn, Poli (2008) presented a method to determine the characteristics of the sampling distribution of a PSO algorithm, and its changes as particles search for better individual best experiences.

## 2.1 Basic algorithm

While the behaviour of the whole system emerges from decentralized local interactions among the particles in the swarm, the individual behaviour of each particle in classic PSO is governed by Eqs. (1) and (2):

$$\begin{cases} v_{ij}^{(t)} = w \cdot v_{ij}^{(t-1)} + \phi_i \cdot \left(pbest_{ij}^{(t-1)} - x_{ij}^{(t-1)}\right) + \phi_s \cdot \left(lbest_{ij}^{(t-1)} - x_{ij}^{(t-1)}\right), \\ x_{ij}^{(t)} = x_{ij}^{(t-1)} + v_{ij}^{(t)}. \end{cases} \quad (1)$$

$$\begin{aligned} \phi_i &= iw \cdot U_{(0,1)} = U_{(0,iw)}, \\ \phi_s &= sw \cdot U_{(0,1)} = U_{(0,sw)}, \\ 0 &\le \left(\phi = \phi_i + \phi_s\right) \le \left(iw + sw\right). \end{aligned} \quad (2)$$

Where:

$v_{ij}^{(t)}$ : $j^{th}$ component of the velocity of particle $i$ at time-step $t$.

$x_{ij}^{(t)}$ : $j^{th}$ coordinate of the position of particle $i$ at time-step $t$.

$\phi_i$ : Individual acceleration coefficient.

$\phi_s$ : Social acceleration coefficient.

$w, iw, sw$ : Inertia, individuality, and sociality weights, respectively.

$U_{(0,a)}$ : Random number from a uniform distribution in the range [0,a] resampled anew every time it is referenced.

$pbest_{ij}^{(t)}$ : $j^{th}$ coordinate of the best position found by particle $i$ by time-step $t$.

$lbest_{ij}^{(t)}$ : $j^{th}$ coordinate of the best position found by any particle in the neighbourhood of particle $i$ by time-step $t$.

The settings of the *inertia* (*w*), *individuality* (*iw*) and *sociality* (*sw*) *weights* in classic PSO governs the individual behaviour of a given particle. Loosely speaking, a high *w* results in higher reluctance to changing the direction of its displacement; a high *iw* results in higher confidence thus typically delaying convergence; and a high *sw* leads to higher conformism thus typically accelerating convergence. There are, however, other issues for different combinations of settings that may modify this behaviour. For instance, increasing individuality over sociality may actually increase convergence speed (see Innocente (2010) or Sienz and Innocente (2010)), while some coefficients result in the particles diverging from rather than clustering around the attractors. One classical means to control the full or even temporary explosions is to limit the size of the updates in every dimension as in Eq. (3).

$$\text{if abs}\left(v_{ij}^{(t)}\right) > v_{\max\ j} > 0 \quad \Rightarrow \quad v_{ij}^{(t)} = \text{sign}\left(v_{ij}^{(t)}\right) \cdot v_{\max\ j} \tag{3}$$

The general flowchart for the whole PSO algorithm is offered in Figure 1, where the update of the particles' velocities and positions are as shown in Eqs. (1) and (2).

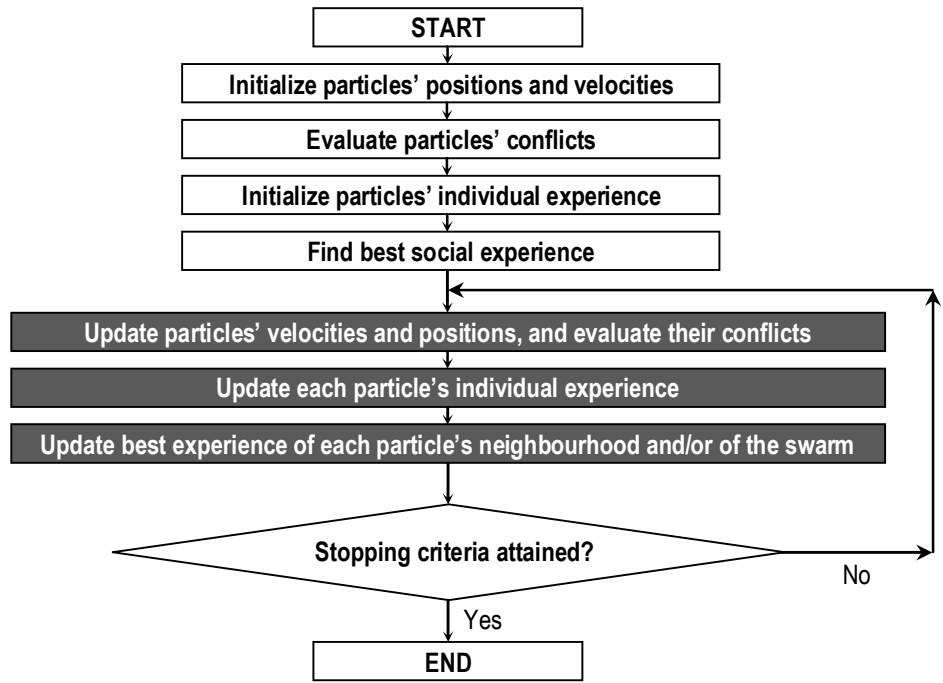

Figure 1: General flowchart for the whole PSO algorithm.

## 2.2 Neighbourhood topology

The speed of convergence of the algorithm as a whole depends on the speed of convergence of each particle towards its attractors (**pbest** and **lbest**) –governed by the coefficients' settings– and also on the speed of spread of information throughout the swarm. The latter is governed by the neighbourhood topology. In other words, the update of **lbest** in Eq. (1), which governs the cooperation between particles, is controlled by the structure of the social network in the swarm. While this topic is beyond the scope of this paper, some popular neighbourhood topologies are presented in Figure 2.

For studies on neighbourhood topologies, refer, for instance, to Kennedy (1998); Kennedy (1999); Suganthan (1999); Mendes (2004); Li (2004); Engelbrecht (2005) (107–109); Kennedy and Mendes (2006); Clerc (2006a) (87–101); Abraham, Liu, and Chang (2006); Mohais (2007); Akat and Gazi (2008); Miranda, Keko, and Duque (2008); and Innocente (2010) (chapter 7).

## 2.3 Additional comments

Another means of affecting the social behaviour is by changing the number of attractors. For instance, by having two social attractors −one local and one global− or by means of the so-called fully-informed PSO in Mendes, Kennedy, and Neves (2004) and Kennedy and Mendes (2006), where every particle is influenced to some extent by all its neighbours.

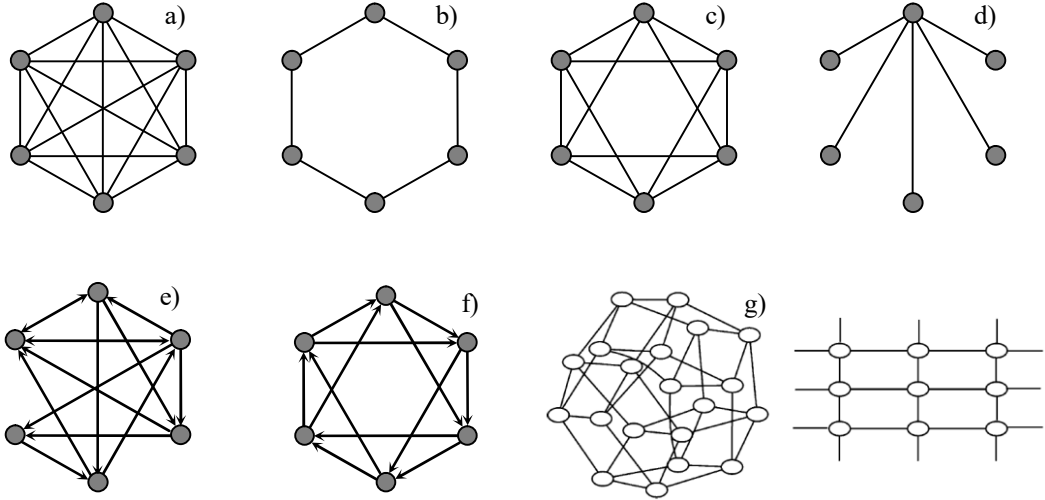

Figure 2: a) *global* topology; b) *ring* topology with two neighbours; c) *ring* topology with four neighbours; d) *wheel* topology; e) *random* topology; f) *forward* topology with two neighbours (from Innocente (2010)); g) *von Neumann* topology (from Kennedy and Mendes (2006)). No arrow means that the link is bidirectional.

## 3  COEFFICIENTS' SETTINGS

The settings of the coefficients $w$, $\phi_i$, and $\phi_s$ in Eq. (1) –and therefore of $\phi$ in Eq. (2)– govern the behaviour of a given particle when pulled by its attractors **pbest** and **lbest**. The first objective should consist of choosing coefficients that ensure convergence of the particle towards its attractors[1]. Once convergence is guaranteed, the form of convergence affects the exploration and exploitation abilities of the particle. This results in controlling the algorithm's capabilities of performing a fine-grain search and of avoiding premature convergence, which are typically conflicting objectives.

### 3.1  Convergence

Since each dimension of the search-space is treated independently from the others in PSO, the theoretical analyses can be performed on one dimensional space. If only one particle is considered, Eq. (1) becomes Eq. (4).

$$\begin{cases} v^{(t)} = w \cdot v^{(t-1)} + \phi_i \cdot \left(pbest^{(t-1)} - x^{(t-1)}\right) + \phi_s \cdot \left(lbest^{(t-1)} - x^{(t-1)}\right), \\ x^{(t)} = x^{(t-1)} + v^{(t)}. \end{cases} \quad (4)$$

A particle '$k$' can be viewed as being pulled by a single attractor ($\mathbf{p}_k$), which results from a randomly weighted average of the components of **pbest**$_k$ and **lbest**$_k$, as shown in Eq. (5). If the coefficients $\phi_i$ and $\phi_s$ were constant for all components of particle '$k$', the attractor $\mathbf{p}_k$ would be located somewhere in the line joining the attractors **pbest**$_k$ and **lbest**$_k$. This is not the case in PSO, although each component of $\mathbf{p}_k$ is indeed a weighted average of the corresponding components of the attractors, as can be observed in Eq. (5).

---

[1] Note that convergence of the full algorithm also depends on the neighbourhood structure.

$$p_{kj} = \frac{\phi_i \cdot pbest_{kj} + \phi_s \cdot lbest_{kj}}{\phi_i + \phi_s} = U_{(pbest_{kj}, lbest_{kj})}. \tag{5}$$

For the one-particle system and the one dimensional case studied within this section, the attractor is given by Eq. (6).

$$p = \frac{\phi_i \cdot pbest + \phi_s \cdot lbest}{\phi_i + \phi_s} = \frac{\phi_i \cdot pbest + \phi_s \cdot lbest}{\phi} = U_{(pbest, lbest)}. \tag{6}$$

Further simplifying the system, consider $\phi_i$ and $\phi_s$ kept constant (randomness removed) and stationary attractors *pbest* and *lbest* (particles' interactions removed). This implies that $\phi$ and the overall attractor $p$ (see Eq. (6)) are also constant. Hence the system in Eq. (4) can be rewritten as in Eq. (7).

$$\begin{cases} v^{(t)} = w \cdot v^{(t-1)} + \phi \cdot (p - x^{(t-1)}), \\ x^{(t)} = x^{(t-1)} + v^{(t)}. \end{cases} \tag{7}$$

Introducing the first equation in Eq. (7) into the second, the second order linear recurrence relation in Eq. (8) can be obtained.

$$x^{(t)} + (\phi - w - 1) \cdot x^{(t-1)} + w \cdot x^{(t-2)} = \phi \cdot p. \tag{8}$$

The two roots ($r_1$ and $r_2$) of the characteristic polynomial of this second order linear recurrence relation are offered in Eq. (9).

$$\begin{aligned} r_1 &= \frac{(1+w)}{2} - \frac{\phi}{2} + \frac{\gamma}{2}, \\ r_2 &= \frac{(1+w)}{2} - \frac{\phi}{2} - \frac{\gamma}{2}, \\ \gamma &= \sqrt{\phi^2 - (2 \cdot w + 2) \cdot \phi + (w-1)^2}. \end{aligned} \tag{9}$$

There are three cases for the general solution of the recurrence relation in Eq. (8), depending on the value of $\gamma$ in Eq. (9):

1) $\gamma^2 > 0$: The roots of the characteristic polynomial are real and different, and the solution is given by Eq. (10).

$$x^{(t)} = p + \frac{r_2 \cdot (p - x^{(0)}) - (p - x^{(1)})}{\gamma} \cdot r_1^t + \frac{-r_1 \cdot (p - x^{(0)}) + (p - x^{(1)})}{\gamma} \cdot r_2^t. \tag{10}$$

2) $\gamma^2 < 0$: The roots of the characteristic polynomial are complex conjugate numbers, and the solution is given by Eq. (11).

$$x^{(t)} = p + \left(\sqrt{w}\right)^t \cdot \left[ -(p - x^{(0)}) \cdot \cos(t \cdot \theta) + \left( \frac{(1 + w - \phi) \cdot (p - x^{(0)}) - 2 \cdot (p - x^{(1)})}{\sqrt{-\gamma^2}} \right) \cdot \sin(t \cdot \theta) \right]. \tag{11}$$

3) $\gamma^2 = 0$: The roots of the characteristic polynomial are real and the same, and the solution is given by Eq. (12).

$$x^{(t)} = p + \left[ -\left(p - x^{(0)}\right) + \left(\left(p - x^{(0)}\right) - \frac{2 \cdot \left(p - x^{(1)}\right)}{1 + w - \phi}\right) \cdot t \right] \cdot \left(\frac{1 + w - \phi}{2}\right)^t. \quad (12)$$

The regions in the '$\phi$–$w$' plane corresponding to each of these three cases are shown in Figure 3 where the points on the parabola correspond to $\gamma^2 = 0$, the region inside the parabola corresponds to the complex conjugates roots ($\gamma^2 < 0$), while the remainder of the plane corresponds to two different and real-valued roots ($\gamma^2 > 0$).

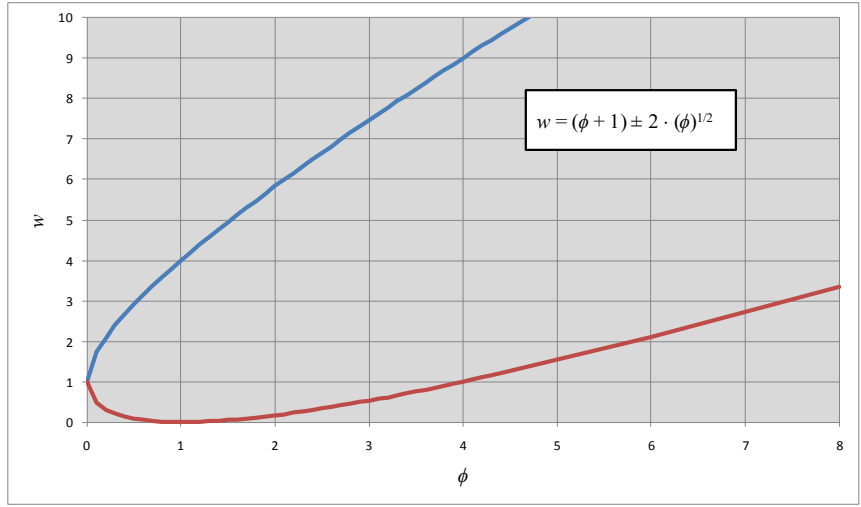

Figure 3: Regions in the '$\phi$–$w$' plane corresponding to each of the three cases of $\gamma$. The points on the parabola correspond to $\gamma^2 = 0$; the region inside the parabola corresponds to the complex conjugates roots; while the remainder of the plane corresponds to two different and real-valued roots.

Therefore, there are two conditions of convergence for this deterministic, isolated particle, each of which is sufficient. In other words, either Eq. (13) or Eq. (14) must be complied with to ensure convergence. Note that these equations are mutually exclusive.

$$\begin{cases} \gamma^2 < 0, \\ w < 1. \end{cases} \quad (13)$$

$$\begin{cases} \gamma^2 \geq 0, \\ -1 < \dfrac{1 + w - \phi \pm \sqrt{\phi^2 - (2 \cdot w + 2) \cdot \phi + (w - 1)^2}}{2} < 1. \end{cases} \quad (14)$$

Solving the inequalities in Eq. (14), and considering the complex region in Figure 3 together with the convergence condition in Eq. (13), the convergence region in the '$\phi$–$w$' plane for the deterministic, isolated particle is the blue-shaded triangle in Figure 4 (refer to Innocente (2010) for further details).

Although there is a region of convergence with $w < 0$, this is of no practical interest because it goes against the concept of inertia. That is to say that the particle tends to keep some momentum from its previous displacement rather than to drastically move in the opposite direction. This can occur due to the attractors, but should not be caused by the inertia component whose function is actually to counteract these drastic changes of direction. Thus the convergence region of practical interest is the area bounded by the four straight lines in Figure 4.

Given that the attractors are stationary, the described behaviour is that of a particle between updates of its attractors. When at least one of them is updated, the particle is driven towards the new *p*. At every stage, convergence of the particle towards the current attractor is ensured. Eventually, better experiences cannot be found, and the particle converges towards the last attractor. Therefore the simplification of considering them stationary does not affect the guaranty of convergence inferred from the blue-shaded triangle in Figure 4.

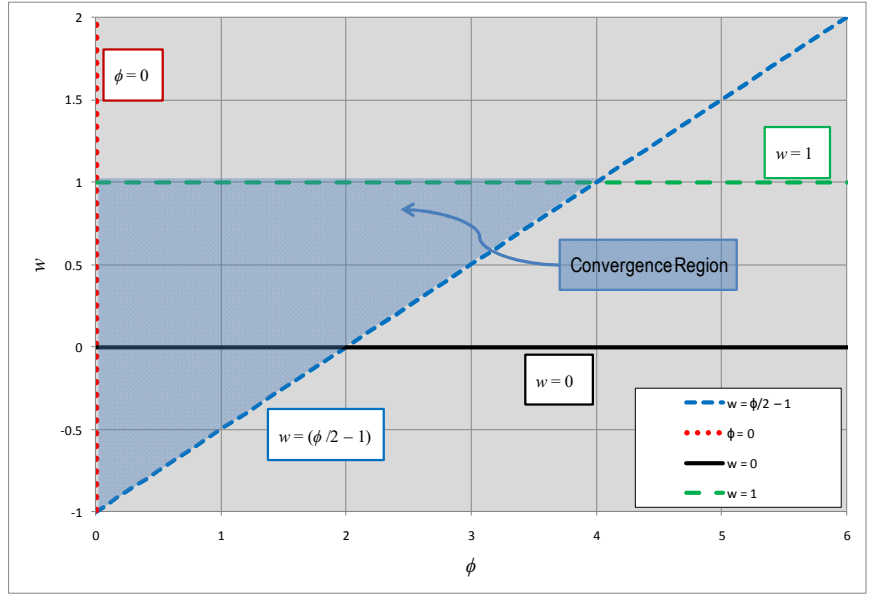

Figure 4: Convergence region in the '$\phi$–$w$' plane for the deterministic, isolated particle (blue-shaded triangle).

Regarding the randomness, if $\phi_{max} = iw + sw$ (see Eq. (2)) is such that the deterministic particle with $\phi = \phi_{max}$ results in convergence, the random particle for which $0 \leq \phi \leq \phi_{max}$ will also converge eventually. It just does so in a more erratic fashion, where some local explosions are possible (refer to Innocente (2010)). The use of the $\mathbf{v}_{max}$ constraint in Eq. (3) typically helps control these so-called 'random explosions' as well as the erratic trajectories.

### 3.2 Speed and form of convergence

Once convergence is ensured, the speed and form of convergence strongly affects the performance of the algorithm, controlling its abilities to avoid premature convergence, to escape local attractors, and to fine-tune the search.

In order to show how different areas within the convergence region result in different speeds and forms of convergence, twenty selected '$\phi$–$w$' pairs are shown in Table 1 and Figure 5. Note that the convergence region within the '$\phi$–$w$' plane in Figure 5 has been trimmed by removing the part corresponding to $w < 0$ (which is of no practical interest).

The twenty '$\phi$–$w$' pairs in Figure 5 have been divided in three groups of eight points each: sub-regions I, II and III. The trajectories of the deterministic particle associated to the points in sub-region I are offered in Figure 6, those associated to the points in sub-region II are shown in Figure 7, while those associated to the points in sub-region III are presented in Figure 8.

As it can be observed, all the 'A' points (A1 to A6) result in either cyclic or pseudo cyclic behaviour. Also refer to Eq. (11) with $w = 1$; Clerc and Kennedy (2002); Innocente (2010); and Sienz and Innocente (2010).

The '$\phi$–$w$' pairs D5 and C6 are on another boundary of the convergence region, where both roots of the characteristic polynomial are real-valued. For the points on that boundary line, the root $r_2 = -1$ in Eq. (9), while the trajectory is governed by Eq. (10). For the points D5 and C6, the other root $-1 < r_1 < 0$ and the trajectories exhibit asymptotic explosions (see Figure 8). Since this second root is greater in magnitude for C6 ($r_1 = -0.50$) than for D5 ($r_1 = -0.25$), the size of the explosion is greater for the former, as shown in Figure 8. The greater the values of $\phi$ and $w$ along that line the greater the explosion (for $w \geq 0$). In fact, '$\phi = 2.00; w = 0.00$' results in cyclic behaviour ($r_1 = 0.00$) whereas '$\phi = 4.00; w = 1.00$' ($r_1 = r_2 = -1$) results in a linear explosion (see Eq. (12)). Greater values lead to exponential explosions with $r_1 = -1$ and $r_2 < -1$.

|   |   | $\phi$ |   |   |   |   |   |
|---|---|---|---|---|---|---|---|
|   |   | 0.50 | 1.00 | 1.50 | 2.00 | 2.50 | 3.00 |
| $w$ | 1.00 | A1 | A2 | A3 | A4 | A5 | A6 |
|   | 0.75 | B1 | B2 | B3 | B4 | B5 | B6 |
|   | 0.50 | C1 | C2 | C3 | C4 | C5 | C6 |
|   | 0.25 | D1 | D2 | D3 | D4 | D5 | D6 |

Table 1: Coordinates of twenty four selected points in the '$\phi$–$w$' plane within or near the convergence region. Three sub-regions containing eight points each are defined (see Figure 5).

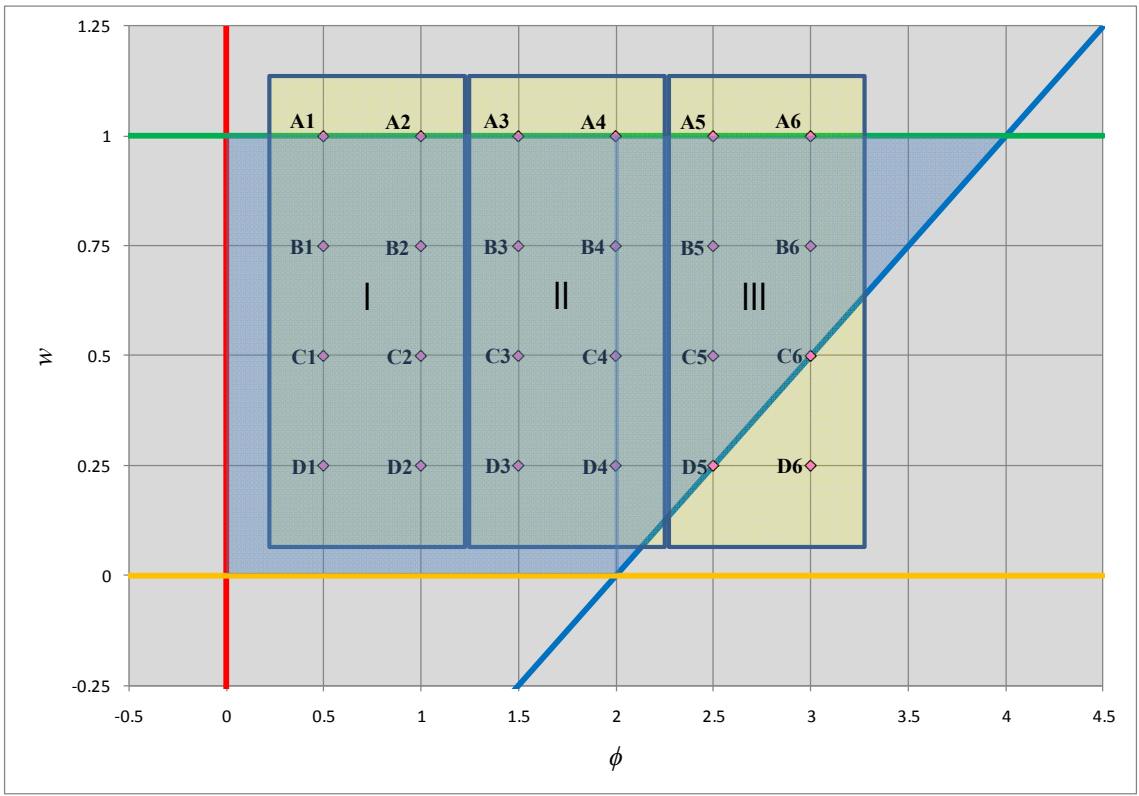

Figure 5: Twenty four selected points in the '$\phi$–$w$' plane within or near the convergence region. Three sub-regions containing eight points each are defined (I, II and III). The points inside the blue-shaded polygon lead to convergence, all 'A' points lead to pseudo cyclic trajectories, whereas C6, D5 and D6 lead to divergence.

For the points along the boundary corresponding to $\phi = 0$, it is self-evident that there would be no movement unless there is an initial velocity. In the latter case, '$\phi = 0; w = 1$' results in a linear explosion (see Eq. (12), as $r_1 = r_2 = 1$).

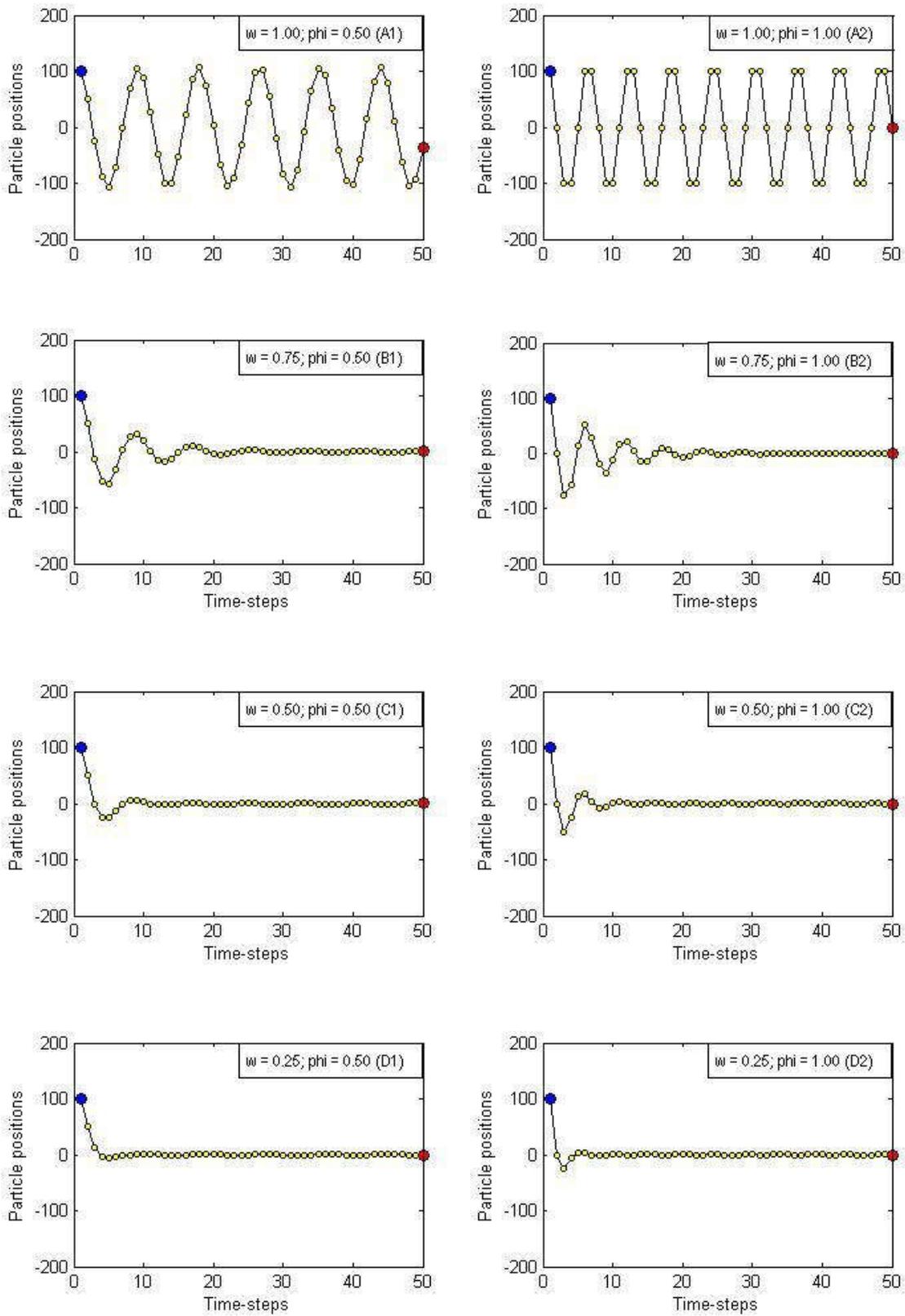

Figure 6: Trajectory of the deterministic particle for the coefficients corresponding to points A1 to D2 within Sub-region I in Figure 5.

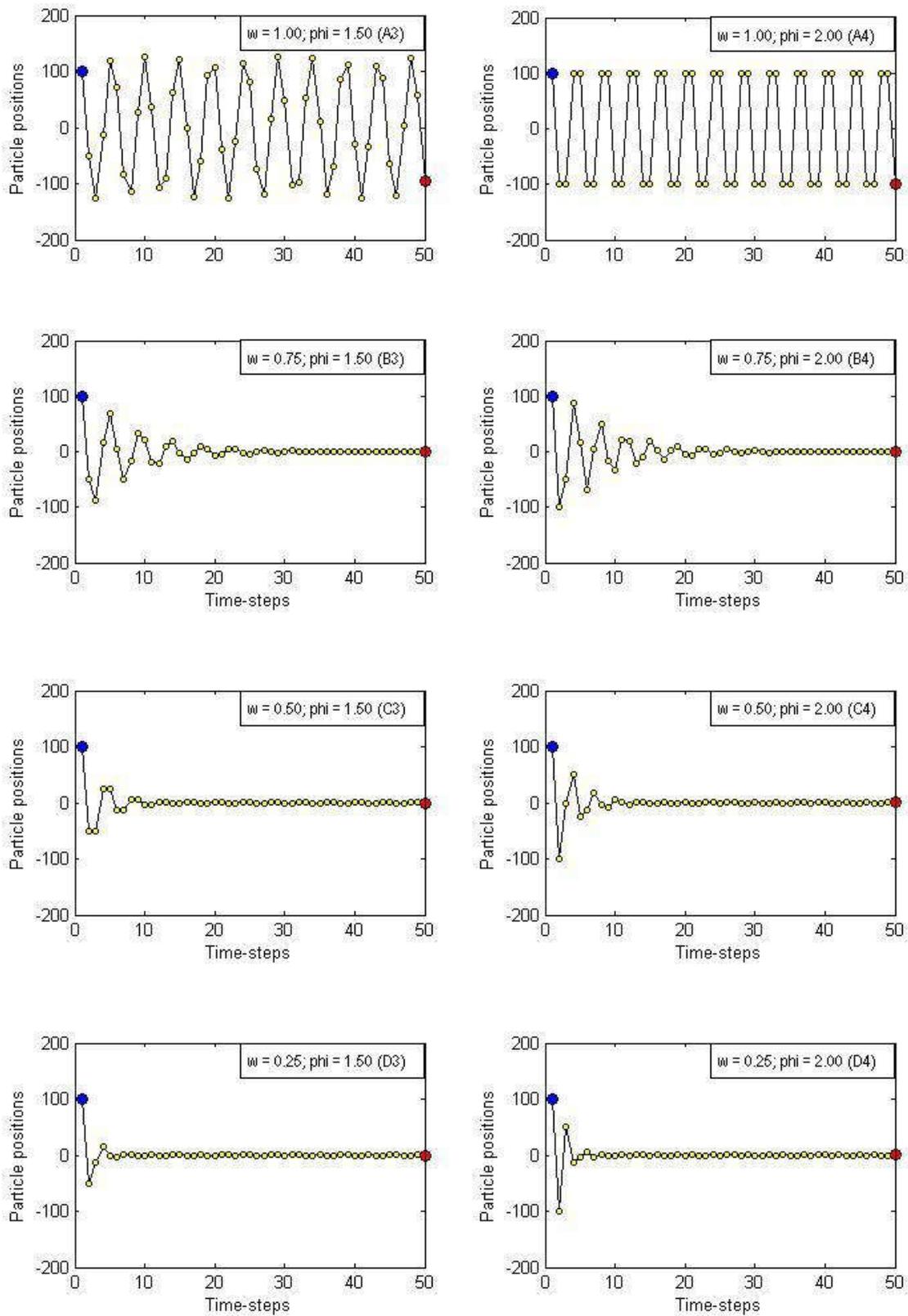

Figure 7: Trajectory of the deterministic particle for the coefficients corresponding to points A3 to D4 within Sub-region II in Figure 5.

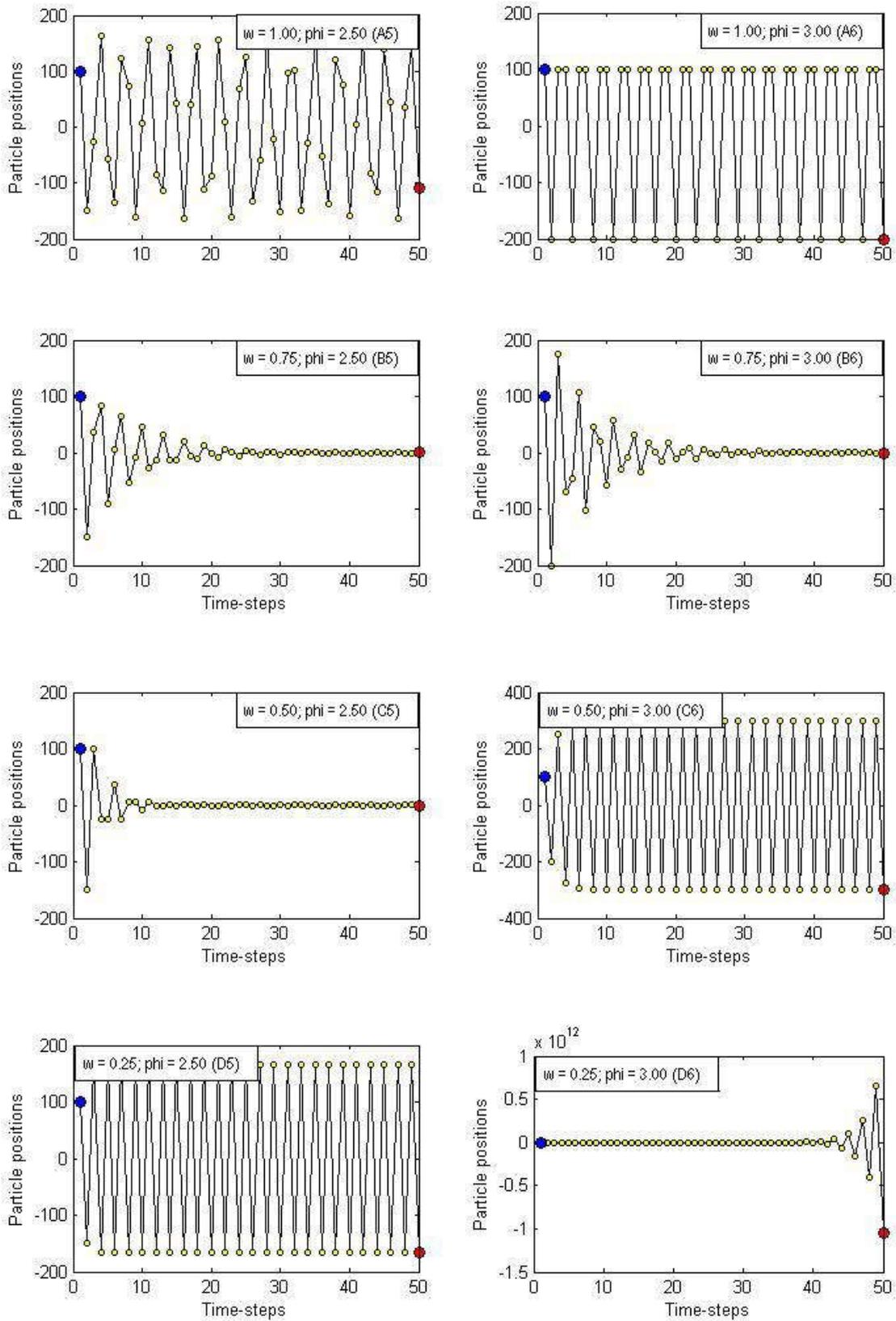

Figure 8: Trajectory of the deterministic particle for the coefficients corresponding to points A5 to D6 within Sub-region III in Figure 5.

In turn, '$\phi=0$; $w=0$' obviously results in no movement, as it can be observed in Eq. (7). Also note that in the latter case $r_1=1$, $r_2=0$, and $\gamma=1$ so that $x^{(t)}=x^{(1)}$ in Eq. (10).

For '$\phi=0$; $0<w<1$', the explosion is asymptotic, where $r_1=1$ and $0<r_2<1$ in Eq. (10).

Note that the trajectories for $\phi=0$ are not included in Figure 6 to Figure 8 because they are of no practical interest.

As to the convergent trajectories, it is obvious that the speed of convergence is a factor to take into account as it allows saving computational cost. However fast convergence is not always desirable as it counteracts the main strength of PSO: its ability to escape poor sub-optimal solutions. Of course a convenient speed of convergence is problem-dependent, where information such as the multimodality of the problem –if known– and the computational resources available need to be taken into account. The slower the convergence the more robust the algorithm is. However, if too slow, convergence may not occur by the time the search is terminated. In addition, the form of convergence is also critical in the performance of the algorithm. Note, for instance, that the speeds of convergence of points B2 and B6 are not too different (refer to Figure 5, Figure 6 and Figure 8), while their forms of convergence are. It is up to the user to choose the type of behaviour desired. In this particular case, for instance, B6 comprises a more robust setting which would be likely to obtain better results in general in abstract search-spaces such as those of mathematical optimization problems. Settings like B2 may be useful when the search-space is in the real world such as in applications like 'swarm robotics', where the sizes of the displacements have a real monetary cost associated. Nonetheless, settings like B5 or B6 should be usually preferred.

### 3.3 Reformulation to control range of randomness

Note that the analysis in the previous sections considers the deterministic particle. Once randomness is reintroduced, average behaviours such as those of B4, B5, C4 and C5 are advisable. However, if $aw=iw+sw=\phi_{max}$ in the classical PSO formulation (see Eq. (2)) is chosen within the convergence region in Figure 5, the $\phi_{mean}$ that can be obtained is restricted. In addition, controlling the range of $\phi$ allows controlling the strength of randomness in the algorithm, where the stronger the influence of randomness the more erratic the trajectories and the more different the actual behaviour is from the average one. A more general formulation is offered in Eq. (15), which allows controlling the strength of randomness.

$$\begin{aligned}
v_{ij}^{(t)} &= w \cdot v_{ij}^{(t-1)} + \phi_i \cdot \left(pbest_{ij}^{(t-1)} - x_{ij}^{(t-1)}\right) + \phi_s \cdot \left(lbest_{ij}^{(t-1)} - x_{ij}^{(t-1)}\right) \\
\\
\phi_i &= ip \cdot \left[\phi_{min} + (\phi_{max} - \phi_{min}) \cdot U_{(0,1)}\right] \\
\phi_s &= sp \cdot \left[\phi_{min} + (\phi_{max} - \phi_{min}) \cdot U_{(0,1)}\right] \\
ip &\in [0,1) \quad ; \quad sp = 1 - ip \\
\\
x_{ij}^{(t)} &= x_{ij}^{(t-1)} + v_{ij}^{(t)}
\end{aligned} \quad (15)$$

### 3.4 Coefficients' settings guidelines

It is advisable that the settings lead to convergence without external mechanisms enforcing it. Thus, at least '$\phi_{mean}$–$w$' should be within the convergence region. Figure 9 shows advised regions within the '$\phi$–$w$' plane from where to choose $w$, $\phi_{min}$ and $\phi_{max}$.

Choose $0.30 \leq w \leq 0.90$. Preferably,

$$0.50 \leq w \leq 0.90 \tag{16}$$

Higher values increase the ability to avoid premature convergence whilst lower values speed up convergence and improve the fine-grain search.

Choose $\phi_{min} \geq 0$ and $2.00 \leq \phi_{max} \leq 4.00$. Advice:

$$\begin{aligned} 0.00 \leq \phi_{min} &\leq 1.00 \\ 2.00 \leq \phi_{max} &\leq 2 \cdot (w+1) \end{aligned} \tag{17}$$

If $\phi_{min} \to 0$, the stochastic acceleration coefficient ($\phi$) may approach zero. Hence a high inertia weight ($w$) with $\phi_{min} \to 0$ may lead to greater local explosions for a sequence of low values of $\phi$ generated. If $\phi_{min} \not\to 0$, the local explosions are more controlled.

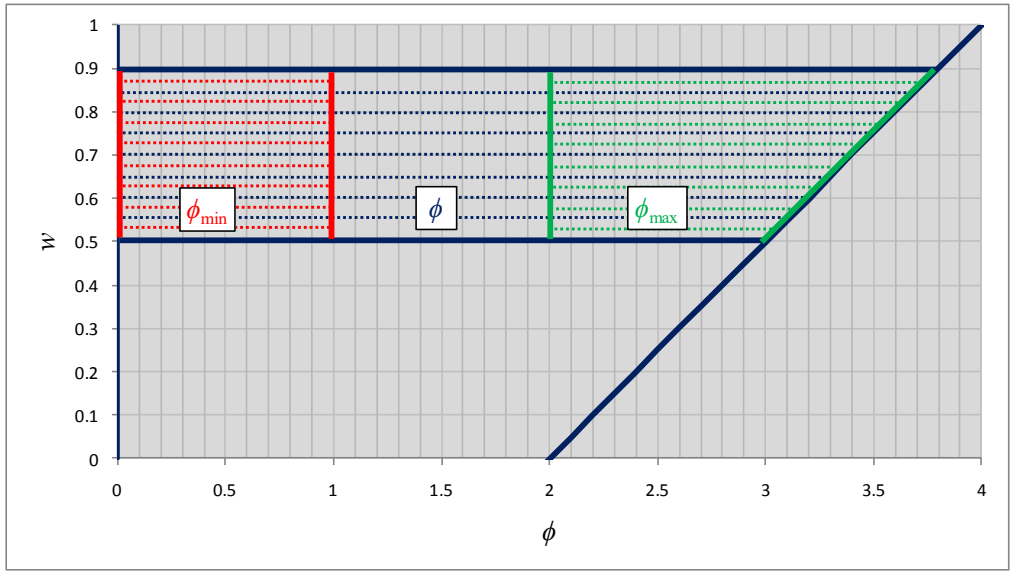

Figure 9: Suggested region in the '$\phi$–$w$' plane from where $\phi$ is to be randomly sampled (blue dotted lines). The regions of suggested upper ($\phi_{max}$) and lower ($\phi_{min}$) limits of $\phi$ are shown in green and red dotted lines, respectively.

Note that $\phi_{min}$ and $\phi_{max}$ define the average behaviour ($\phi_{mean}$) as well as the strength awarded to randomness. For the same average behaviour, a greater interval of $\phi$ results in higher exploration, more erratic behaviour, and slower convergence. Advice:

$$1.00 < \phi_{mean} = 0.50 \cdot (\phi_{max} + \phi_{min}) < 2.00 \tag{18}$$

Note that lower accelerations lead to higher amplitudes and lower frequencies of the oscillatory trajectories around the attractors. Higher amplitudes widen the exploration region while higher frequencies result in the particles overflying their attractors a higher number of times, and approaching them from both sides in each dimension (advisable).

Choose $ip$, $sp$, and $v_{max}$. Advice:

$$ip = sp = 0.50 \tag{19}$$

$$v_{max\,j} = 0.50 \cdot (x_{j\,max} - x_{j\,min}) \tag{20}$$

As mentioned before, greater values of the coefficients are more robust. In the absence of any information regarding the problem, general-purpose settings that would work reasonably well on most problems are $0.70 \leq w \leq 0.80$ and $\phi_{max}$ close to the convergence boundary.

**Classical PSO formulation**

To translate the proposed formulation into the classical one, replace $\phi_{min}$ in Eq. (17) by Eq. (21). Other relations between the two formulations are offered in Eq. (22).

$$\boxed{\phi_{min} = 0} \tag{21}$$

$$\left.\begin{matrix} iw = ip \cdot \phi_{max} \\ sw = sp \cdot \phi_{max} \end{matrix}\right\} \Rightarrow \begin{cases} \phi_i = iw \cdot U_{(0,1)} = U_{(0,iw)} \\ \phi_s = sw \cdot U_{(0,1)} = U_{(0,sw)} \end{cases} \tag{22}$$

Given Eq. (21), higher values of $\phi_{max}$ also have the indirect effect of increasing the effect of randomness (widen the range of $\phi$). That is, the lower the $\phi_{max}$ the more similar the actual behaviour is to the average behaviour. And therefore, higher values of $\phi_{max}$ indirectly decrease the speed of convergence and result in more erratic behaviour.

**Constricted PSO formulation**

Choose $aw$ and $0 < \kappa < 1$. Advice:

$$\boxed{\begin{matrix} aw > 4 \text{ (slightly)} \\ \kappa \rightarrow 1 \end{matrix}} \tag{23}$$

Replace Eqs. (16) and (17) by Eq. (24).

$$\boxed{\begin{matrix} w = \begin{cases} \dfrac{2 \cdot \kappa}{aw - 2 + \sqrt{aw^2 - 4 \cdot aw}} & \text{if } aw \geq 4 \\ \kappa & \text{otherwise} \end{cases} \\ \phi_{min} = 0 \\ \phi_{max} = w \cdot aw \end{matrix}} \tag{24}$$

## 4 FINAL REMARKS AND FUTURE RESEARCH

A formal analysis of the influence of the coefficients in the velocity update equation on the trajectory of a deterministic particle was presented, and the sets of settings that result in the convergence of the deterministic particle towards its attractors were presented. Within this set, the types of behaviours to be expected for different combinations of settings were provided, allowing the user to choose the type of behaviour desired for a given problem and available resources. The classical PSO algorithm was reformulated to allow better control of the strength of randomness desired, and guidelines were provided for the coefficients' settings. The relations between the proposed formulation and both the classical and the constricted PSO (the latter from (Clerc and Kennedy, 2002)) were offered, so that the guidelines can also be applied to them if desired.

The next step in our research is the study of the influence of randomness on the trajectory of the isolated random particle, some of which can be found in (Innocente, 2010). Advanced studies including the influence of randomness can be found in (Jiang et al., 2007), (Clerc, 2006b), and (Poli, 2008).

To complete the reincorporation of the complexity of the full paradigm to the simplified system studied, the interaction between particles and the influence of varying the relative strength of individuality and sociality need to be studied. These aspects have been analyzed to some extent in (Innocente, 2010) and (Sienz and Innocente, 2010). Finally, while these studies help the user understand the behaviour of the system and select the coefficients that result in the desired behaviour for a given problem and resources, how to decide on what behaviour should be desired is not always straightforward.